\documentclass[letterpaper]{article} 
\usepackage{aaai2027}
\usepackage{times}  
\usepackage{helvet}  
\usepackage{courier}  
\usepackage[hyphens]{url}  
\usepackage{graphicx} 
\usepackage{natbib}  
\usepackage{caption} 
\usepackage{amsmath}
\usepackage{amssymb}
\usepackage{algorithm} 
\usepackage{booktabs}
\usepackage{mathtools}
\usepackage{array}
\usepackage{multirow}
\usepackage{multicol}
\usepackage{subcaption}
\usepackage{enumitem}
\usepackage{soul}
\usepackage{bm}
\usepackage{pifont}
\newcommand{\cmark}{\ding{51}}
\usepackage{url}
\usepackage{amsthm}

\usepackage[noabbrev]{cleveref}
\usepackage{amsfonts}

\usepackage[autostyle]{csquotes}  
\usepackage{algpseudocode} 
\usepackage{mwe}
\usepackage{physics}
\usepackage{xcolor}
\usepackage{tikz}
\usetikzlibrary{positioning,arrows.meta,fit}

\usepackage{newfloat}
\usepackage{listings}
\DeclareCaptionStyle{ruled}{labelfont=normalfont,labelsep=colon,strut=off} 
\floatstyle{ruled}
\newfloat{listing}{tb}{lst}{}
\floatname{listing}{Listing}
\title{Topological Neural Dynamics: A Neuron-wise Framework for Sequence Modeling}
\author{
    Borui Cai\textsuperscript{\rm 1},
    Yao Zhao\textsuperscript{\rm 2}
}
\affiliations{
    \textsuperscript{\rm 1}Hangzhou International Innovation Institute, Beihang University, China\\
    \textsuperscript{\rm 2}Victoria University, Melbourne, Australia\\
    caibr@buaa.edu.cn, yao.zhao@vu.edu.au
}

\usepackage{bibentry}

\begin{document}

\maketitle

\begin{abstract}
Most existing sequence models, such as RNNs, LSTMs, and Transformers, share a common structural principle: layer-wise dynamics, where all neurons in the same layer co-evolve through a shared parameterized operator, leaving individual neurons no freedom to evolve independently. Yet in many complex dynamical systems, rich global behavior emerges precisely from locally evolving units interacting through structured connectivity. Inspired by this principle, we introduce Topological Neural Dynamics (TND), a sequence modeling framework that shifts computation from layer-wise to neuron-wise dynamics. TND represents a neural system as a directed neuron graph, an interaction operator, and a local dynamics function, where each neuron evolves independently and collective computation emerges from interactions through the explicit graph topology. We instantiate TND as a discrete-time graph-coupled dynamical system and evaluate it as a case study on a behavior cloning task in single-player Pong. Compared with Vanilla RNN, Sparse RNN, LSTM, S4, Closed-form continuous-time neural network (CfC), and Transformer baselines, TND achieves the best catch rate and a mean of 17.47 consecutive catches per round, more than three times that of the strongest baseline. These results suggest that shifting from layer-wise to neuron-wise dynamics provides an effective inductive bias for sequence modeling.
\end{abstract}

\begin{links}
    {\link{Code}{https://github.com/brcai/tnd_pong}}
\end{links}


\section{Introduction}
Sequence modeling is central to many problems in artificial intelligence, including robotic control \cite{robot}, autonomous driving \cite{autonomous}, and game playing \cite{imitation}. In these settings, a model must map a stream of observations to a sequence of actions while preserving the temporal dependencies that govern the underlying process. For example, a behavior cloning model \cite{behavior} must reproduce expert behavior from demonstrations, a robotic controller must generate temporally coherent actions from evolving sensory inputs, and an autonomous system must maintain performance under changing state observations. The key challenge is not merely to process individual inputs, but to learn how information should evolve, interact, and persist over time.

\par
\begin{figure}[t]
\centering
\includegraphics[width=0.8\linewidth]{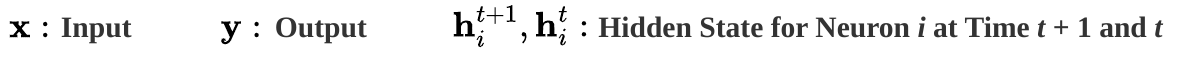}
\begin{subfigure}{0.48\linewidth}
    \centering
    \includegraphics[width=\linewidth]{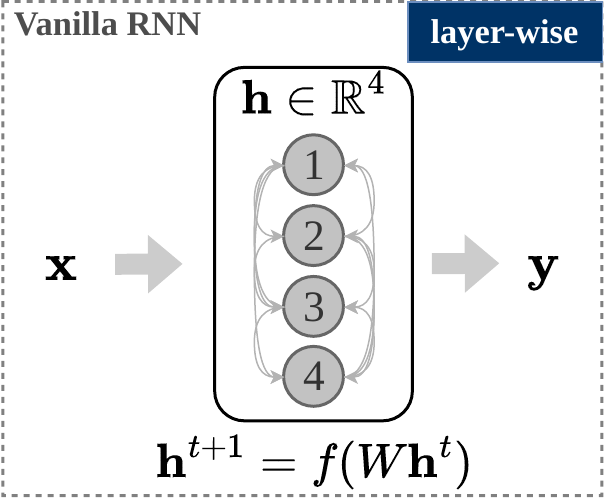}
    \caption{Vanilla RNN}
    \label{fig:existing}
\end{subfigure}
\begin{subfigure}{0.48\linewidth}
    \centering
    \includegraphics[width=\linewidth]{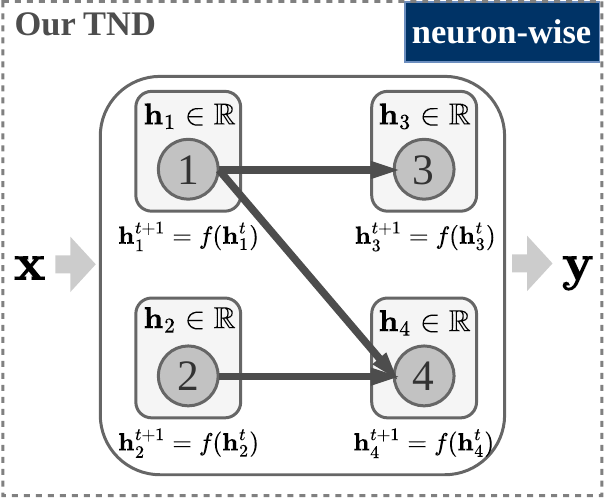}
    \caption{Our TND}
    \label{fig:tdn}
\end{subfigure}
\caption{Comparison of Vanilla RNN and TND. 
Vanilla RNN updates a monolithic hidden state $\mathbf{h}\in\mathbb{R}^4$ through a shared transformation $W$ by function $f(\cdot)$, coupling the temporal evolution of all neurons. In contrast, TND assigns each neuron an independent hidden state and allows information exchange only through explicit graph connections.}
\label{fig:intro}
\vspace{-1 em}
\end{figure}

 A dominant approach to this challenge is \textit{neural sequence modeling}, where a parameterized model learns to map observation sequences to action sequences~\cite{seq}. Recurrent Neural Network (RNN) and gated variants, such as Long Short-Term Memory (LSTM)~\cite{lstm}, maintain hidden states that are recurrently updated over time, enabling historical information to influence future predictions. More recent continuous-time models~\cite{ctrnn} extend this idea by modeling state evolution through continuous-time dynamics. Although these architectures differ in how they parameterize temporal evolution, they share a common structural principle, i.e., \textbf{layer-wise} dynamics: as the example of vanilla RNN shown in Fig.~\ref{fig:intro}(a), the state of a layer is updated through a shared parameterized transformation applied to the entire hidden representation. As a result, the evolution of each neuron is not governed by its own autonomous dynamics, but is coupled with the evolution of other neurons through the same layer-level operator. This layer-wise coupling forces neurons to co-evolve as components of a single global state, restricting their ability to follow heterogeneous, unit-specific trajectories. This limits the model’s structural flexibility, particularly when the target process requires diverse local dynamics and adaptive self-organization.

\par In fact, many real-world complex dynamical systems achieve rich global behavior precisely because their individual units evolve according to local dynamics and self-organize through interactions~\cite{complex1,complex2,nature1}. For example, biological neurons follow their own membrane dynamics and interact through synaptic connectivity, collectively giving rise to coordinated neural computation~\cite{giannari2022model}. Similarly, in epidemic systems, individuals undergo local infection-recovery dynamics, while population-level spreading patterns emerge through contact networks~\cite{pastor2015epidemic}. In these systems, global behavior is not imposed by a single shared evolution rule over all units; rather, it emerges from the interaction and self-organization of locally evolving components. Conventional sequence models largely suppress this mechanism by enforcing shared layer-wise evolution, thereby reducing the independent degrees of freedom available to individual neurons. This mismatch limits their expressive capacity for modeling systems in which heterogeneous local dynamics and emergent collective behavior are essential.

\par To address this limitation, we introduce \emph{Topological Neural Dynamics} (TND), a \textbf{neuron-wise} evolution framework that models each neuron as an autonomous dynamical unit. As illustrated in Fig.~\ref{fig:intro}(b), rather than updating all neurons through a shared layer-level operator, TND allows each neuron to evolve according to its own local dynamics, while a directed neuron graph explicitly defines how information flows among neurons. Through these topological interactions, neurons can influence one another without losing their individual degrees of freedom, enabling collective computation to emerge from the self-organization of locally evolving units. This design shifts the computational granularity from layer-wise dynamics, where a monolithic state vector is jointly transformed by a shared operator, to neuron-wise dynamics, where heterogeneous neuron trajectories interact through an explicit topology. By doing so, TND provides a more flexible modeling paradigm for sequential computation with heterogeneous local dynamics.


\par Formally, we define TND as a neural system represented by a triple $\mathcal{T}=(\mathcal{G},\mathcal{I},\mathcal{F})$, where (1) $\mathcal{G}$ is a directed neuron graph that separately represents neurons as nodes and interaction topology as edges; (2) $\mathcal{I}$ is a neuron interaction operator that propagates and transforms signals along graph edges; and (3) $\mathcal{F}$ is a neuron dynamics function governing the internal evolution of each neuron. Through this design, TND shifts computation from a shared layer-wise operator to independent neuron-wise dynamics, where each neuron is free to evolve according to its own local state, and collective behavior emerges from their self-organization through the explicit graph topology. In addition, this design exposes neuron dynamics and interaction topology as two independent and composable design dimensions, enabling each to be specified, analyzed, and modified independently of the other. Note that Spiking Neural Networks (SNNs) \cite{snn} can be viewed as a special case of TND, where $\mathcal{F}$ is instantiated as leaky integrate-and-fire dynamics with a threshold firing function.

\par We instantiate TND as a graph-coupled discrete neural system with a randomly generated interaction topology and evaluate it on a behavior cloning task. Specifically, the model learns to control a paddle in a single-player Pong game by imitating human demonstrations, using ball and paddle positions as input and predicting human-like actions. We compare TND against six sequence-modeling baselines with comparable parameter counts: Vanilla RNN, Sparse RNN, LSTM, S4 \cite{s4}, Closed-form Continuous-time Neural Network (CfC) \cite{cfc}, and Transformer. The results show that TND achieves the best ball catch rate, with a mean of 17.47 consecutive catches per round, compared with 6.14 for the strongest baseline (CfC). This demonstrates that the model flexibility introduced by neuron-wise dynamics substantially improves such behavioral sequence learning. Our contributions are:
\begin{itemize}[leftmargin=*, noitemsep]
    \item We propose TND, a framework that shifts neural computation from layer-wise dynamics to neuron-wise dynamics, where each neuron evolves independently and collective behavior emerges from self-organization through an explicit graph topology.
    \item We present an instantiation of TND as a discrete-time dynamical system on a sparse neuron graph, and show that TND outperforms representative baselines on a behavior cloning task, achieving more than three times the mean consecutive catches of the strongest baseline.
\end{itemize}

\section{Related Work}
\subsection{Sequence modeling}
RNN \cite{rnn} models temporal dependencies through a hidden state updated by a learned transition operator. Later variants, including LSTM \cite{lstm} and GRU \cite{gru}, adopt gating mechanisms to integrate historical information and relieve gradient vanishing. Another line is the continuous-time extensions, such as Continuous-time Recurrent Neural Network (CTRNN) \cite{ctrnn}, which define neuron dynamics directly as ODE with a fixed leaky-integrator structure. Then, neural ODE \cite{ode} generalizes this by parameterizing the ODE with a neural network, enabling more expressive continuous-time dynamics trained via the adjoint method. Liquid Time-constant Network \cite{liquid} and CfC \cite{cfc} build further on the CTRNN formulation with input-dependent time constants and analytical ODE approximations, respectively.
State space models, including S4 \cite{s4} and Mamba \cite{mamba}, achieve competitive sequence modeling through structured linear recurrence. Transformer \cite{transformer} forgoes recurrent state entirely in favor of global self-attention, but lacks an explicit temporal state, making it less suited for continuous control tasks that require persistent memory across timesteps. \emph{Across all these approaches, neurons are forced to co-evolve through a shared operator, leaving individual neurons limited freedom to evolve independently.}

\subsection{Topology in neural computation}
Some existing works have provided early research on the topology for neural network performance. 
Reservoir computing approaches, such as echo state networks \cite{echo} demonstrate that a fixed random recurrent reservoir with only a trained linear readout can support rich temporal computation. Liquid state machines \cite{lsm} extend this principle to spiking neural networks, showing that biologically plausible random connectivity similarly enables complex spatiotemporal computation.
A parallel line of work investigates learning or pruning connectivity. Sparse recurrent training methods \cite{srnn} show that sparse weight matrices in recurrent networks improve generalization, and adopt pruning and re-growing of edges during the dynamic exploration of optimal sparse connectivities \cite{sparsernn1, sparsernn2}. Neural architecture search approaches \cite{nas} treat connectivity as a discrete variable to be optimized alongside weights.

Recent advances in connectomics provide strong biological evidence that neuron dynamics and interaction topology are independent components of neural computation. For example, models constrained solely by the complete wiring diagram of an adult Drosophila brain~\cite{drosophila} accurately reproduce sensorimotor processing~\cite{fly} and predict neural activity at single-neuron resolution~\cite{flyvisual}. Similarly, in the human brain, structural connectome topology has been shown to constrain the propagation of fast neural dynamics across brain regions, independent of the local dynamics of individual neurons \cite{human}.
This biological principle directly motivates TND, where each neuron is endowed with its own dynamics and neurons self-organize through an explicit graph topology.

\begin{figure}[tp]
\centering
\includegraphics[width=1\linewidth]{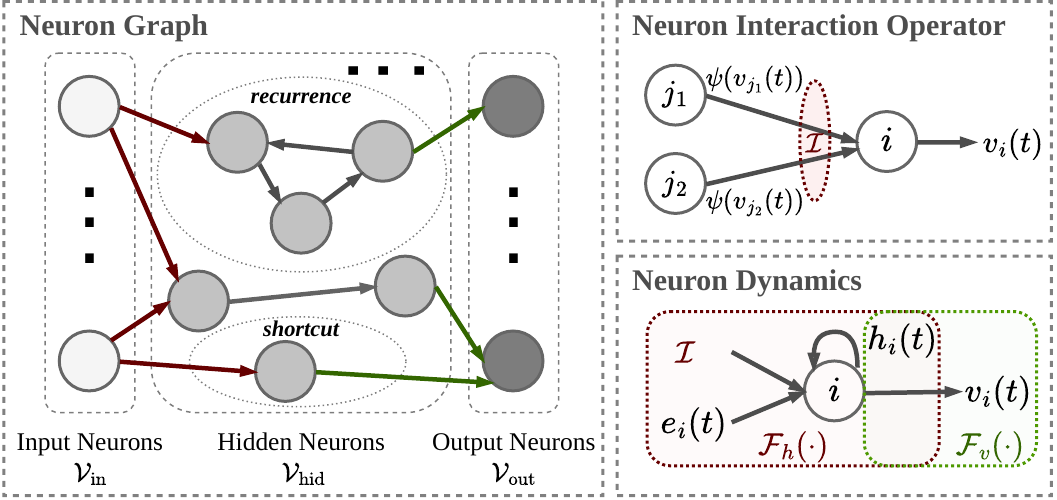}
\caption{Overview of the proposed TND framework.}
\label{fig:top}
\end{figure}

\section{Topological Neural Dynamics}
Topological Neural Dynamics (TND) is a framework that shifts neural computation from layer-wise dynamics to neuron-wise dynamics, where each neuron evolves independently and collective behavior emerges from self-organization through an explicit graph topology. It represents a neural network as a neuron graph, where nodes instantiate single-neuron dynamics and edges define topological interactions among neurons. Formally, TND is defined as
\begin{equation}
    \mathcal{T} = (\mathcal{G}, \mathcal{I}, \mathcal{F})
\end{equation}
where $\mathcal{G}$ is the \textbf{neuron graph} (i.e., interaction topology), $\mathcal{I}$ is the \textbf{neuron interaction operator} defined over the edges of $\mathcal{G}$, and $\mathcal{F}$ is the internal \textbf{neuron dynamics}. Same as conventional sequential computation model, given an input sequence $\boldsymbol{x}(t)$, TND produces an output sequence $\boldsymbol{y}(t)$:
\begin{equation}
    \boldsymbol{y}(t)=\mathcal{T}(\boldsymbol{x}(t))
\end{equation}

\subsection{Technical Details}
Each component in TND is detailed below.

\smallskip
\noindent\textbf{Neuron Graph} \bm{$\mathcal{G}$}. The neuron graph is a directed graph
$\mathcal{G}=(\mathcal{V},\mathcal{E})$, where $\mathcal{V}$ is the set of neurons and $\mathcal{E}\subseteq\mathcal{V}\times\mathcal{V}$ is the set of directed edges. Each neuron $i\in\mathcal{V}$ receives an input signal $\boldsymbol{u}_i(t)$ and produces an output $v_i(t)$. The neuron set is partitioned into input, hidden, and output neurons:
\begin{equation}
    \mathcal{V}=\mathcal{V}_{\rm in}\cup\mathcal{V}_{\rm hid}\cup\mathcal{V}_{\rm out}
\end{equation}
Input neurons $i\in\mathcal{V}_{\rm in}$ receive external signals from the system input, output neurons $i\in\mathcal{V}_{\rm out}$ produce the system output, and hidden neurons $i\in\mathcal{V}_{\rm hid}$ are internal computational units not directly observed outside the system.
Note that such a graph model also enables flexible design of neuron connectivity, i.e., introducing patterns like recurrence or short-cut throughout the topology. 

\smallskip
\noindent\textbf{Neuron Interaction Operator} \bm{$\mathcal{I}$}. The neuron interaction operator $\mathcal{I}$ aggregates incoming signals for each neuron according to the graph topology $\mathcal{G}$. For neuron $i\in\mathcal{V}$, its in-neighborhood is
\begin{equation}
    \mathcal{N}_{\mathcal{G}}(i)=\{j\in\mathcal{V}:(j,i)\in\mathcal{E}\}
\end{equation}
where $(j,i)$ denotes a directed edge from neuron $j$ to neuron $i$. Given the neuron output $\boldsymbol{v}(t)$, the interaction signal received by neuron $i$ is
\begin{equation}
\label{eq:interaction signal}
    \mathcal{I}(i,\ \boldsymbol{v}(t)) = \{ \psi(v_{j}(t)) : j\in \mathcal{N}_\mathcal{G}(i) \} 
\end{equation}
where $\psi(\cdot)$ transforms each incoming neighbor output. $\mathcal{I}(i,\boldsymbol{v}(t))$ collects all transformed incoming neighbor outputs as the neuron input. 

\smallskip
\noindent\textbf{Neuron Dynamics} \bm{$\mathcal{F}$}. The neuron dynamics $\mathcal{F}=(\mathcal{F}_h,\mathcal{F}_v)$ consists of a state-evolution function $\mathcal{F}_h$ and an output function $\mathcal{F}_v$. Each neuron $i\in\mathcal{V}$ maintains a hidden state $h_i(t)$ and produces an output $v_i(t)$:
\begin{align}
    \frac{d h_i(t)}{dt}&=\mathcal{F}_h\!\left(h_i(t),\mathcal{I}(i,\boldsymbol{v}(t)),\boldsymbol{e}_i(t)\right) \\
    v_i(t)&=\mathcal{F}_v\!\left(h_i(t)\right)
\end{align}
where $\boldsymbol{e}_i(t)$ is the external input injected into neuron $i$. For input neurons $i\in\mathcal{V}_{\rm in}$, $\boldsymbol{e}_i(t)$ is assigned from the system input $\boldsymbol{x}(t)$; for hidden and output neurons, $\boldsymbol{e}_i(t)=\boldsymbol{0}$. The system output is formed by collecting the outputs of output neurons:
\begin{equation}
\label{eq:system output}
    \boldsymbol{y}(t)=\{v_i(t):i\in\mathcal{V}_{\rm out}\}
\end{equation}
The specific forms of $\mathcal{F}_h$, $\mathcal{F}_v$, and $\psi(\cdot)$ depend on the chosen TND instantiation.

\subsection{Discrete-Time Formulation}
The preceding formulation defines TND as a general dynamical framework. To support practical sequence modeling with discretely observed inputs, we instantiate TND in discrete time. Given 
$\mathcal{T}=(\mathcal{G},\mathcal{I},\mathcal{F})$, the system evolves over timesteps 
$t=\{1,\ldots,T\}$. At each $t$, neuron interaction operator $\mathcal{I}$ aggregates transformed 
outputs from the previous timestep by Eq.~(\ref{eq:interaction signal}). Each neuron then updates its hidden state and output as
\begin{align}
h_i^{t+1}&=\mathcal{F}_h\!\left(h_i^{t},\mathcal{I}(i,\boldsymbol{v}^{t}),\boldsymbol{e}_i^{t+1}\right), \\
v_i^{t+1}&=\mathcal{F}_v(h_i^{t+1})
\end{align}
\noindent where $\mathcal{I}(i,\boldsymbol{v}^{t})=\{ \psi(v_{j}^{t}) : j\in \mathcal{N}_\mathcal{G}(i)\}$. $h_i^t$ and $v_i^t$ are the hidden state and output of neuron $i$ at timestep $t$, respectively. The initial states and outputs $\{h_i^0,v_i^0\}_{i\in\mathcal{V}}$ are set to zero or randomly initialized. Given an input sequence
\begin{equation}
X=\{\boldsymbol{x}^1,\boldsymbol{x}^2,\ldots,\boldsymbol{x}^T\},\quad\boldsymbol{x}^t\in\mathbb{R}^{|\mathcal{V}_{\rm in}|}
\end{equation}
TND produces an output sequence
\begin{equation}
Y=\{\boldsymbol{y}^1,\boldsymbol{y}^2,\ldots,\boldsymbol{y}^T\},\quad\boldsymbol{y}^t\in\mathbb{R}^{|\mathcal{V}_{\rm out}|}
\end{equation}
where the system output is obtained by collecting the outputs of output neurons by Eq.~(\ref{eq:system output}). Thus, the discrete-time TND system maps the current input and previous neural state to the current output:
\begin{equation}
\boldsymbol{y}^{t+1}=\mathcal{T}(\boldsymbol{h}^{t},\boldsymbol{x}^{t+1})
\end{equation}

\begin{figure}[tp]
\centering
\includegraphics[width=0.8\linewidth]{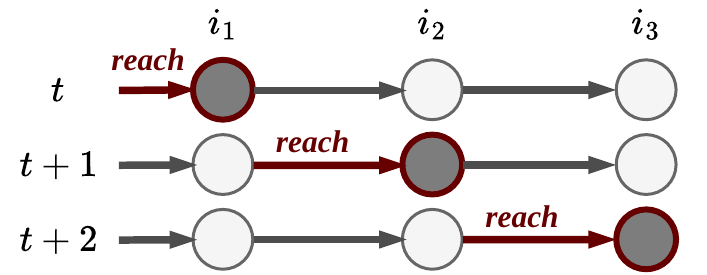}
\caption{Signal propagation delay.}
\label{fig:delay}
\end{figure}

\noindent\textbf{Note.} TND introduces a topology-induced propagation delay, where signals require multiple timesteps to travel from input neurons to distant parts of the neuron graph, as shown in Fig.~\ref{fig:delay}. Unlike existing sequence models that usually rely on explicit gates, shared recurrent updates, or learnable time constants to control temporal behavior, TND obtains multi-scale temporal processing directly from graph topology. Because different input-output paths can have different lengths, neurons naturally operate at different temporal scales: neurons close to the input respond quickly to external changes, while distant neurons integrate information over longer horizons. Since output neurons receive only information that has propagated through the graph, TND is implicitly encouraged to retain past inputs and predict future states from stored history. This memory-prediction behavior is supported by two complementary mechanisms: unit-level memory, where each neuron maintains its own local hidden state, and network-level memory, where recurrent connectivity allows information to circulate and persist across timesteps. We show the effectiveness of such a design in the Experimental Results section.

\begin{table*}[tp]
\centering
\small
\caption{Model Performance on Pong Game}
\begin{tabular}{l | ccccccccc}
\toprule
\multirow{2.5}{*}{\textbf{Methods}} & \multicolumn{3}{c}{\textbf{$l=20$}}& \multicolumn{3}{c}{\textbf{$l=40$}} & \multicolumn{3}{c}{\textbf{$l=60$}}\\
\cmidrule(lr){2-4}  \cmidrule(lr){5-7} \cmidrule(lr){8-10}
&$\mathrm{Rate}$ &$\mathrm{Mean}$ &$\mathrm{Max}$ &$\mathrm{Rate}$ &$\mathrm{Mean}$ &$\mathrm{Max}$ &$\mathrm{Rate}$ &$\mathrm{Mean}$ &$\mathrm{Max}$ \\
\midrule
Vanilla RNN    &0.63	&1.69	&10	&0.61	&1.57   &13	 &0.55   &1.2	 &18\\
Sparse RNN     &0.74	&2.81	&17	&0.78	&3.47   &18	 &0.75   &3.07	 &16\\
LSTM           &0.83	&4.78	&27	&0.85	&5.64   &26	 &0.77   &3.32   &27\\
S4             &0.69    &2.21   &19 &0.67   &2.01   &20  &0.65   &1.85   &16\\
Transformer    &0.52	&1.06	&8	&0.48	&0.93   &4	 &0.42   &0.72	 &4\\
CfC            &0.86	&6.14	&46	&0.84	&5.41   &53	 &0.84   &5.25	 &35\\
\textbf{TND}   &\textbf{0.94}	&\textbf{14.81}	&\textbf{72}	&\textbf{0.95}	&\textbf{17.47}  &\textbf{68}	 &\textbf{0.95}   &\textbf{17.29}	 &\textbf{72}\\
\bottomrule
\end{tabular}
\label{tab:main}
\vspace{-1 em}
\end{table*}

\section{Experiments}
\label{sec:Experiments}
We evaluate TND on a single-player Pong game under a behavior cloning setup.

\subsection{Experimental Setup}
\noindent\textbf{TND Instantiation}. 
In this experiment, we use a simple discrete-time instantiation of TND, intentionally keeping the neuron dynamics and aggregation scheme minimal. This design choice isolates the effect of neuron-wise dynamics as a structural principle, so that performance gain can mainly be attributed to the neuron-wise design itself rather than implementation complexity. Specifically, neuron graph $\mathcal{G}$ is constructed with a fixed random topology: each candidate edge is independently sampled with probability $p$, which we refer to as the connection density. Recurrence and shortcuts naturally appear in such an instantiation, which is essential for sequential computation. Output neurons are restricted to receive signals only from hidden neurons. The neuron interaction operator $\mathcal{I}$ applies a incoming signals aggregation:
\begin{equation}
    \mathcal{I}(i,\boldsymbol{v}^{t})=\sum_{j\in\mathcal{N}_{\mathcal{G}}(i)}W_{ij}v_j^{t}
\end{equation}
where $W_{ij}$ is the learnable interaction weight on edge $(j,i)\in\mathcal{E}$. The neuron dynamics function $\mathcal{F}$ updates each hidden state using a leaky-integrator rule:
\begin{align*}
    h_{i}^{t+1} =& (1-\tau)h_{i}^{t} + \tau(tanh(\Sigma_{j\in \mathcal{N}_{\mathcal{G}}(i)} W_{ij}v_{j}^{t} + \\ 
    & w^{in}_{i}e_{i}^{t+1} + b_{i} + \alpha_{i}h_{i}^{t} )) \\
v_{i}^{t+1} =& tanh( h_{i}^{t+1} )
\end{align*}
where $e_i^{t+1}$ is the external input to neuron $i$ at timestep $t+1$ and is nonzero only for input neurons. The learnable parameters are the edge weights $\{W_{ij}\}_{(j,i)\in\mathcal{E}}$ for $\mathcal{I}$ and the neuron-specific parameters $\{w_i^{\rm in}, b_i, \alpha_i\}_{i\in\mathcal{V}}$ for $\mathcal{F}$. Here, $w_i^{\rm in}$ controls the strength of external input, $b_i$ is the neuron bias, $\alpha_i$ controls self-state feedback, and $\tau$ is a fixed integration step size. Spectral normalization is applied to the initial weight matrix $W$ to improve dynamical stability.

\smallskip
\noindent\textbf{Data Preparation}. Following the behavior cloning setting, we use human gameplay demonstrations as training data for single-player Pong. During demonstration collection, the human player controls the paddle to catch the ball and completes the game without failure. The demonstration data forms a paired sequence $(X,Y)$, where each input $\boldsymbol{x}^{t}\in X$ records the ball and paddle positions, i.e., $\{x_{\rm ball},y_{\rm ball},x_{\rm paddle},y_{\rm paddle}\}$, and each label $\boldsymbol{y}^{t}\in Y$ records the corresponding human action from \{\emph{left}, \emph{right}, \emph{stay}\}. We quantize each input $\boldsymbol{x}^{t}$ into a 24-dimensional binary vector and encode each action $\boldsymbol{y}^{t}$ as a 3-dimensional one-hot vector. In total, 20,000 continuous input--action pairs are collected for model training.

\smallskip
\noindent\textbf{Evaluation Metric}. 
Each model is evaluated by its ability to control the paddle and catch the ball. Although the training objective does not explicitly specify ``catching the ball'', this behavior is implicitly encoded in the human demonstrations, where the player successfully catches all balls without failure. During evaluation, each model controls the paddle for 10 sessions. In each session, the ball position is randomly initialized and the game runs for 2,000 steps. If the model misses the ball, the current round terminates and a new round starts with another randomly initialized ball position. We report the overall catch rate $\mathrm{Rate}$ (\%):
\begin{equation}
    \mathrm{Rate}
    =
    \frac{C_{\rm succ}}{C_{\rm succ}+C_{\rm fail}}
\end{equation}
where $C_{\rm succ}$ and $C_{\rm fail}$ denote the total numbers of successful catches and failures, respectively. We also report the mean and maximum consecutive catches per round, denoted by $\mathrm{Mean}$ and $\mathrm{Max}$. Since Pong has a small state space, a model may occasionally enter a limit-cycle attractor, where the ball and paddle follow a periodic trajectory and the overall catch rate becomes trivially $100\%$. Such trials are excluded because they do not reflect the model's imitation ability.

\begin{figure*}[t]
\centering
\subfloat[Vanilla RNN]{\includegraphics[width=1.2in]{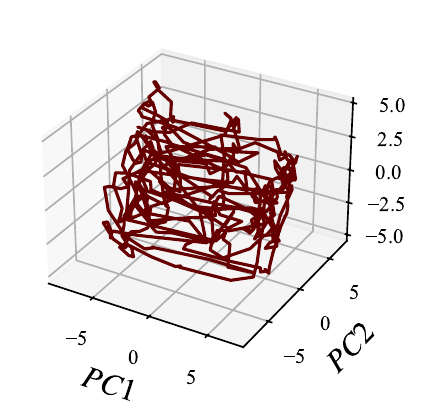}}
\hspace{-1 em}
\subfloat[Sparse RNN]{\includegraphics[width=1.2in]{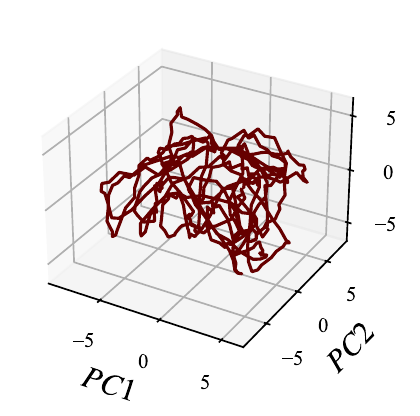}}
\hspace{-1 em}
\subfloat[LSTM]{\includegraphics[width=1.2in]{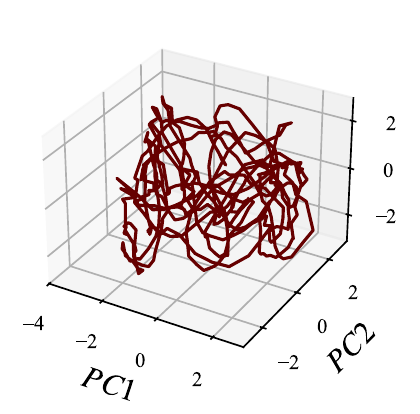}}
\hspace{-1 em}
\subfloat[S4]{\includegraphics[width=1.2in]{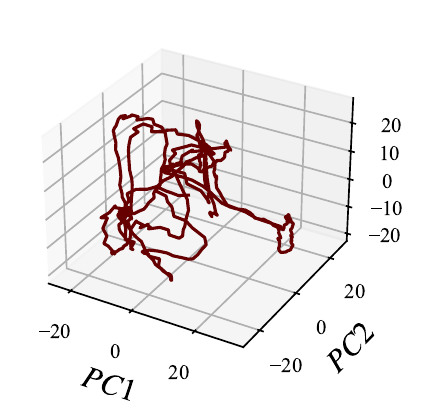}}
\hspace{-1 em}
\subfloat[CfC]{\includegraphics[width=1.2in]{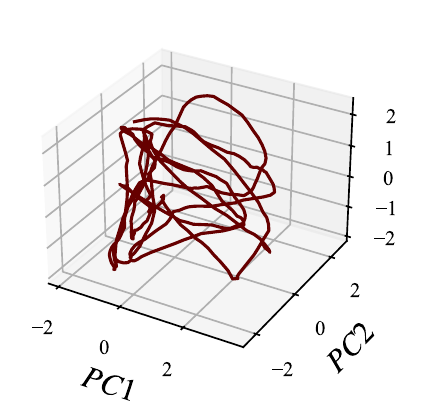}} 
\hspace{-1 em}
\subfloat[TND]{\includegraphics[width=1.2in]{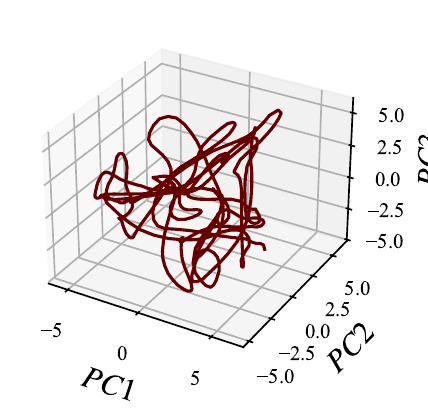}}
\caption{3D trajectory of model hidden states during game playing, obtained by Principal Component Analysis (PCA), where PC1, PC2, and PC3 mean the first three principal components obtained by PCA.}
\label{fig:trajectory}
\end{figure*}

\smallskip
\noindent\textbf{Baselines}. We compare TND with representative sequence-modeling baselines: Vanilla RNN, Sparse RNN, LSTM, Transformer, S4 and CfC. Vanilla RNN and LSTM represent standard recurrent models; Sparse RNN tests whether sparse connectivity alone explains TND's performance; Transformer provides an attention-based non-recurrent baseline with sinusoidal positional encoding; S4 serves as a structured state-space model baseline that captures long-range dependencies via state-space dynamics; and CfC serves as a continuous-time neural dynamics baseline with closed-form updates. Together, these baselines cover the major approaches to sequential modeling, allowing us to evaluate the benefits of neuron-wise dynamics against diverse architectural designs.

\smallskip
\noindent\textbf{Training Implementation}. 
We construct training samples using sliding windows of length $l$, where each sample is 
$\{(\boldsymbol{x}^{t-l},\ldots,\boldsymbol{x}^{t}),\boldsymbol{y}^{t}\}$. 
All models take a 24-dimensional quantized ball--paddle location vector as input and output a 3-dimensional action vector. Specifically, $\hat{\boldsymbol{y}}^{t}=\mathrm{Model}(\boldsymbol{x}^{t-l},\ldots,\boldsymbol{x}^{t})$, and the training loss is the mean squared error, i.e., $\mathcal{L}=\sum_{t=1}^{T}\left\|\boldsymbol{y}^{t}-\hat{\boldsymbol{y}}^{t}\right\|^{2}$. For Vanilla RNN, Sparse RNN, LSTM, Transformer, and S4, the hidden size is searched over $\{64,128,256,512\}$ and the number of layers over $\{1,2,3,4,5\}$. For CfC and TND, the number of neurons is searched over $\{200,400,600,800\}$ to obtain comparable parameter counts. The elapsed-time input in CfC is searched over $\{0.01,0.1,1\}$. The sparsity factor $p$ in TND and Sparse RNN is searched over $\{0.2,0.4,0.6,0.8\}$. The integration step size $\tau$ in TND is searched over $\{0.2,0.4,0.6,0.8\}$. All experiments are implemented in Python using PyTorch and run on a Linux platform with an NVIDIA A30 GPU.

\subsection{Experimental Results}
\noindent\textbf{Main Results}. 
We evaluate all models with input window lengths $l \in \{20,40,60\}$, and summarize the results in Table~\ref{tab:main}. TND consistently achieves the best performance across all input lengths and evaluation metrics. For example, with an input length of 40, TND achieves a catch rate of 0.95, a mean of 17.47 consecutive catches per round, and a maximum of 68 consecutive catches in a single round.

\smallskip
\noindent\underline{\textbf{\emph{Takeaway}}}. Several observations can be made.
\begin{itemize}[leftmargin=*, noitemsep]
    \item TND substantially outperforms all baselines, especially in mean consecutive catches. This indicates stronger long-horizon behavioral consistency rather than only improved single-step accuracy.
    \item TND remains robust across input lengths, suggesting that its graph-based connectivity provides effective temporal information pathways. In contrast, RNNs and LSTM degrade when the input length increases from 40 to 60, which indicates difficulty in exploiting longer temporal contexts.
    \item Baseline comparisons highlight the importance of TND's neuron-wise dynamics. Transformer performs poorly across all settings, which suggests that attention without a persistent recurrent state is less suitable for this task. Sparse RNN improves upon Vanilla RNN, showing that connectivity structure matters, but its performance remains far below TND, indicating that sparsity alone is insufficient.
\end{itemize}

\begin{figure*}[t]
\centering
\includegraphics[width=2.2in]{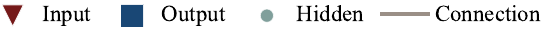}
\vspace{-0.5 em}
\subfloat[$p=0.2\ \rm (Rate=0.93)$]{\includegraphics[width=1.7in]{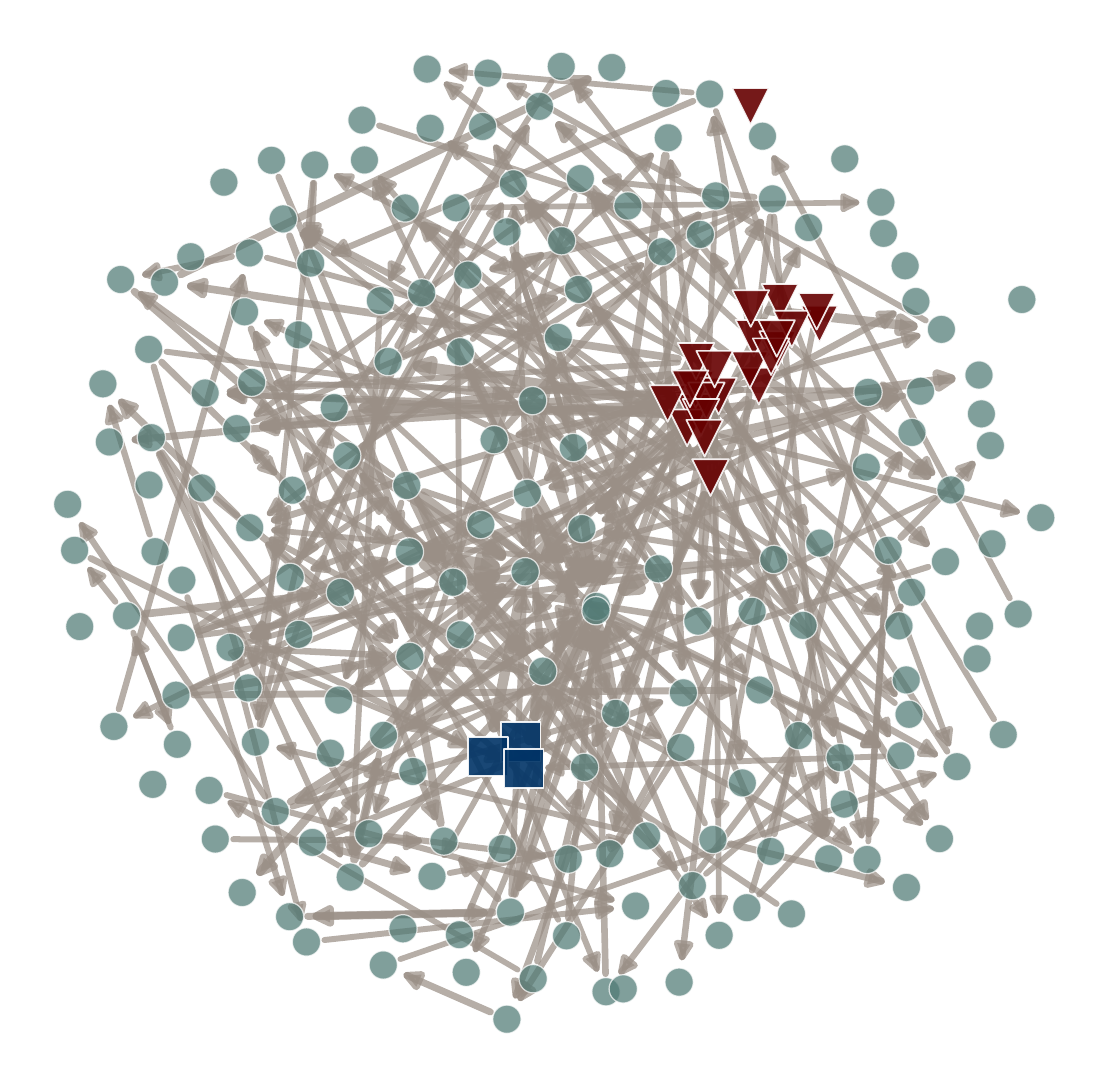}}
\subfloat[$p=0.4\ \rm (Rate=0.94)$]{\includegraphics[width=1.7in]{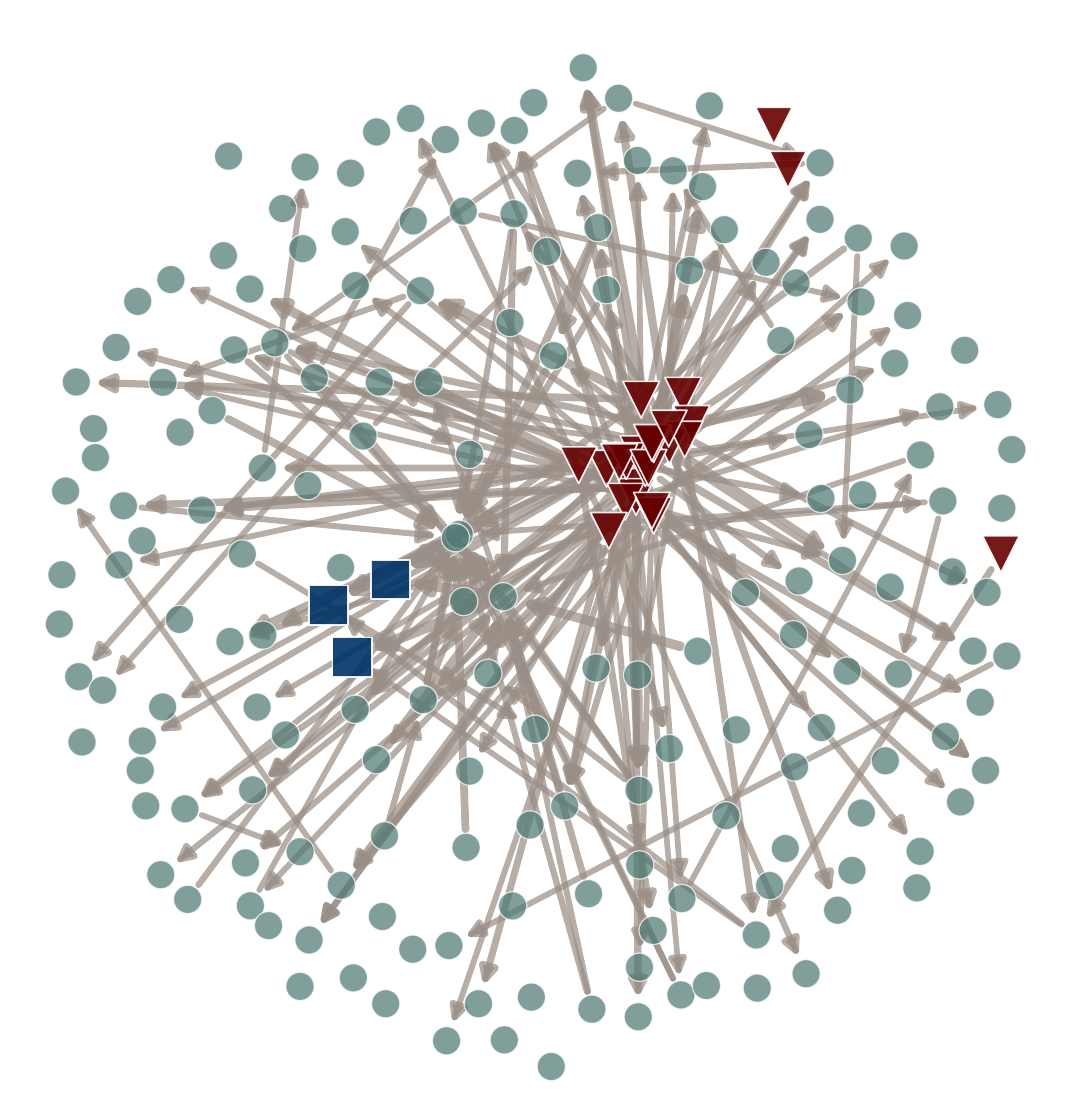}} 
\subfloat[$p=0.6\ \rm (Rate=0.91)$]{\includegraphics[width=1.7in]{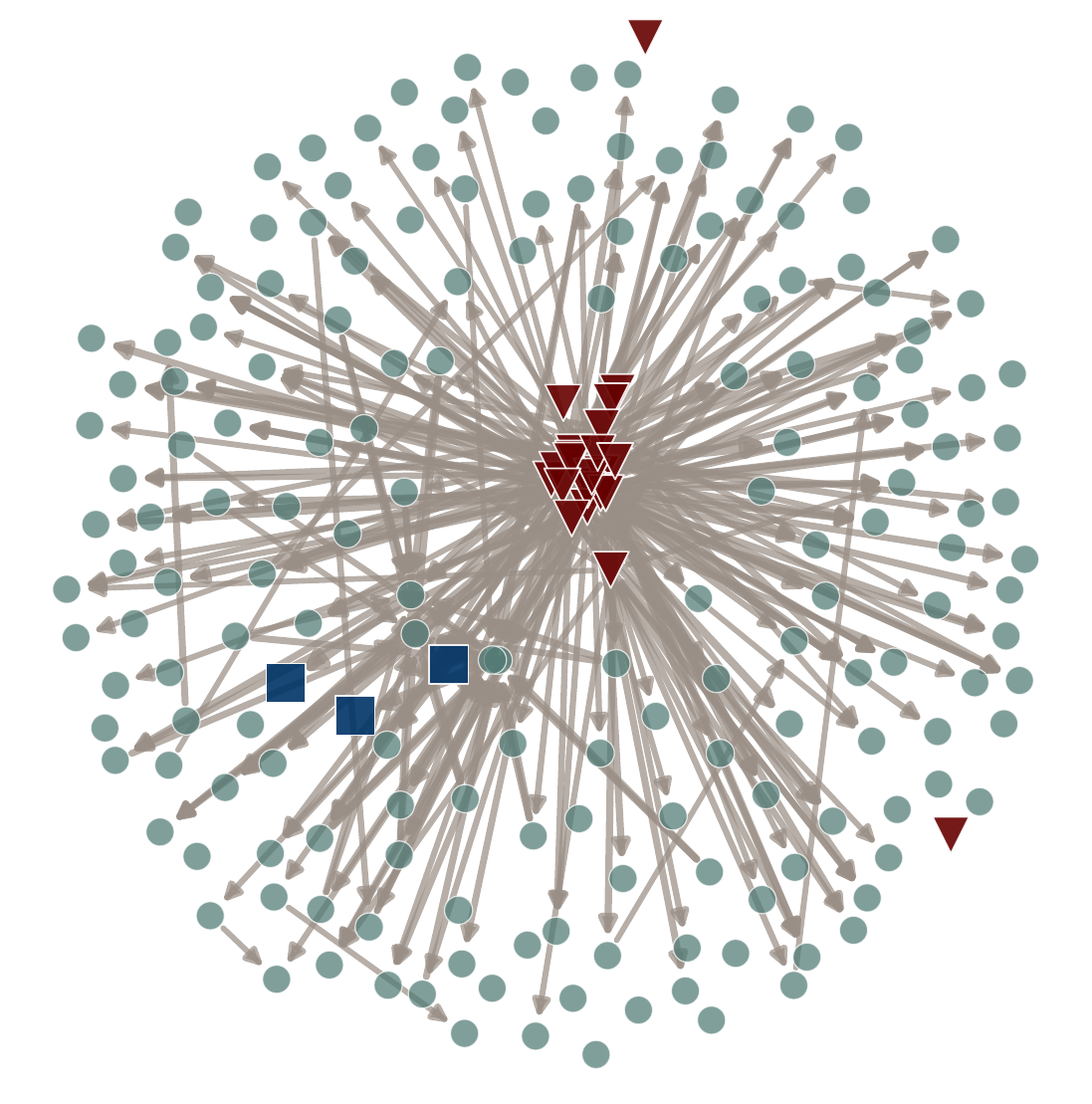}}
\subfloat[$p=0.8\ \rm (Rate=0.88)$]{\includegraphics[width=1.7in]{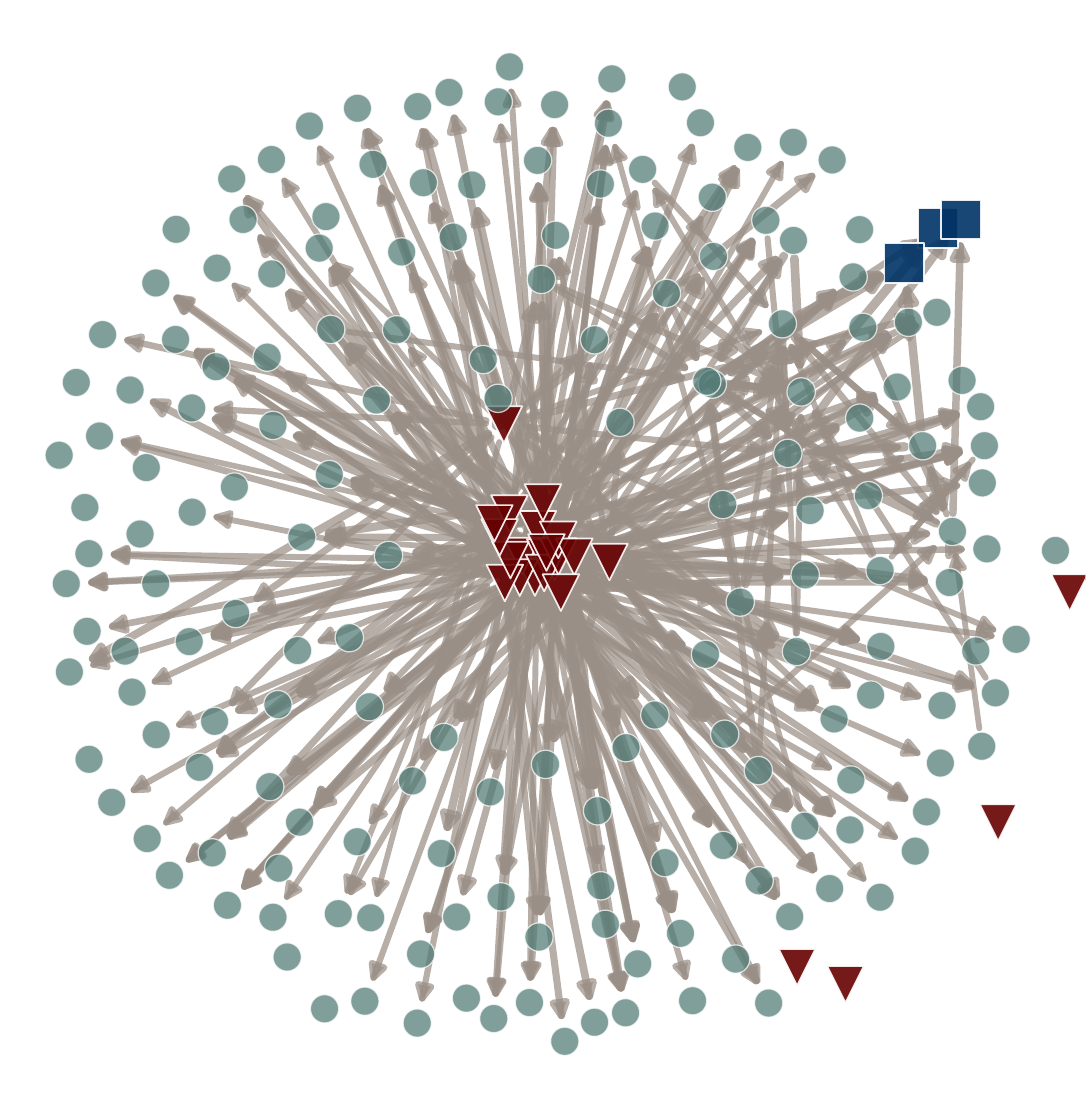}} \\
\caption{TND topology visualization, only strong edges with weights $|w|>0.3$ are shown.}
\label{fig:topology}
\vspace{-1 em}
\end{figure*}

\begin{figure*}[t]
\centering
\subfloat[$p=0.2\ \rm (Rate=0.92)$]{\includegraphics[width=1.7in]{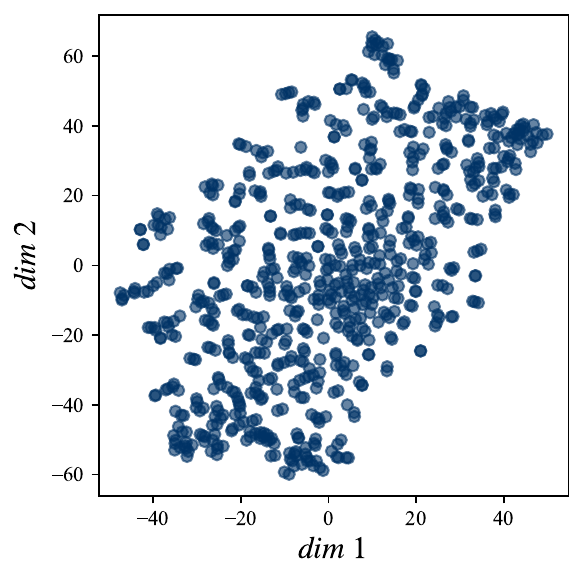}}
\subfloat[$p=0.4\ \rm (Rate=0.92)$]{\includegraphics[width=1.7in]{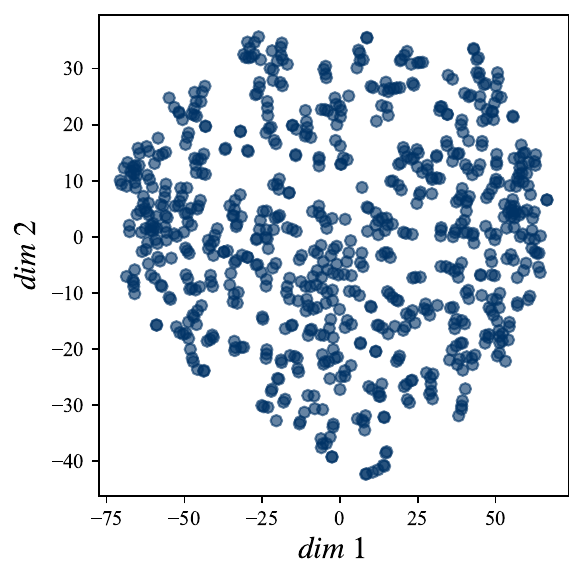}} 
\subfloat[$p=0.6\ \rm (Rate=0.88)$]{\includegraphics[width=1.7in]{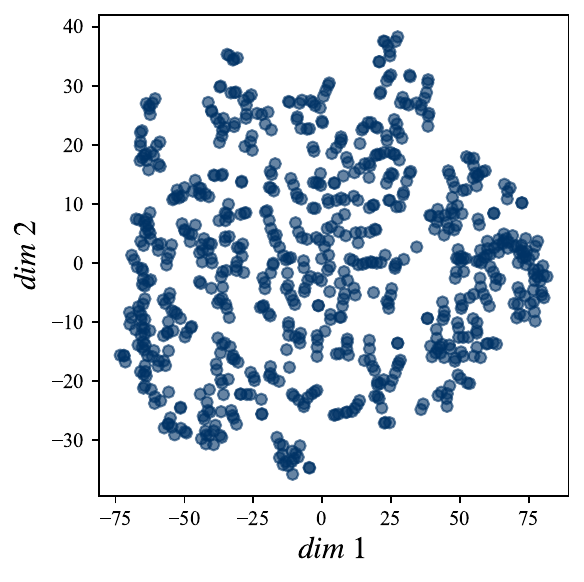}}
\subfloat[$p=0.8\ \rm (Rate=0.60)$]{\includegraphics[width=1.7in]{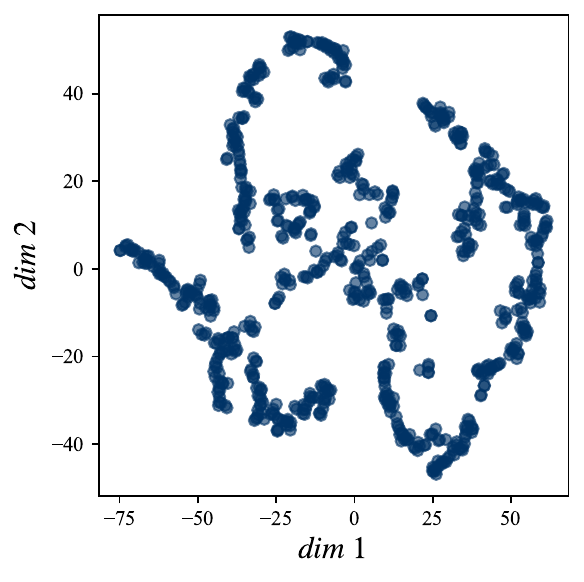}} \\
\caption{TND neuron activity analysis. Each neuron's 500-timestep hidden-state trajectory is projected into 2D via t-SNE, where neurons with similar temporal activity patterns appear close together.}
\label{fig:act}
\vspace{-1 em}
\end{figure*}

\noindent\textbf{Trajectory Analysis}. Fig.~\ref{fig:trajectory} visualizes the hidden-state trajectories of each model during a game-play session, projected onto the top three PCA components. Vanilla RNN and Sparse RNN exhibit frequent sharp transitions in the PCA space, indicating abrupt changes in their hidden states. LSTM and S4 show more confined trajectories, but still contain noticeable transitions. CfC also exhibits sharp changes despite its continuous-time formulation. In contrast, TND produces markedly smoother and more structured trajectories, suggesting more gradual and coherent hidden-state evolution.

\smallskip
\noindent\underline{\textbf{\emph{Takeaways}}}. Takeaways are summarized as follows.
\begin{itemize}
    \item Baselines show sharper global state updates. The sharp transitions observed in Vanilla RNN, Sparse RNN, LSTM, S4, and CfC suggest that their hidden representations are more sensitive to instantaneous input changes. Although LSTM uses gating and CfC adopts a continuous-time formulation, both still rely on globally coupled hidden-state updates. This may lead to abrupt changes in the internal state and less stable sequential control behavior.
    \item TND exhibits smoother internal dynamics. The smooth PCA trajectory indicates that TND's hidden state evolves gradually rather than changing abruptly between consecutive timesteps. This behavior is consistent with TND's neuron-wise design: each neuron maintains an independent local hidden state, and signals propagate through the directed neuron graph over time. As a result, external inputs do not immediately perturb the entire hidden representation, leading to more coherent global state evolution.
\end{itemize}


\noindent\textbf{Model Size Analysis}. Table~\ref{tab:mp} compares the number of trainable parameters for each model under its best-performing hyperparameter configuration. TND contains 0.36M parameters, which is comparable to Vanilla RNN (V-RNN, 0.28M) and Sparse RNN (S-RNN, 0.34M), but far smaller than LSTM (3.20M), Transformer (Trans., 6.34M), S4 (3.14M), and CfC (1.50M).
This indicates that TND's performance improvement is not due to a larger parameter budget. Instead, it suggests that the gain mainly comes from its neuron-wise dynamics and graph-based propagation structure, which provide a more parameter-efficient mechanism for sequential control.

\begin{table}[t]
\small
\centering
\caption{Comparison of Model Parameters (in million)}
\setlength{\tabcolsep}{3pt}
\begin{tabular}{c|ccccccc}
\toprule
Model & V-RNN & S-RNN &LSTM &Trans. &S4 &CfC & TND\\
\midrule
Para. & 0.28 &0.34	&3.20	   &6.34	&3.14 &1.50	&0.36 \\
\bottomrule
\end{tabular}
\label{tab:mp}
\end{table}

\smallskip
\noindent\underline{\textbf{\emph{Takeaway}}}. TND achieves strong performance with a parameter size close to simple recurrent baselines, demonstrating that its advantage comes from structural design rather than model scale.

\smallskip
\noindent\textbf{TND Topology Analysis}. 
To understand why TND outperforms the baselines, we analyze its learned topology and hidden activity under different connection densities $p$. We use TND with 200 neurons, while fixing $\tau=0.8$. Fig.~\ref{fig:topology} visualizes the adjacency graphs of trained TND models. Only strong edges with weights magnitude greater than 0.3 are shown. The learned structures vary clearly with $p$. In dense TND ($p=0.8$), strong edges are mainly concentrated between input and hidden neurons, which indicates more direct input-to-hidden transmission. In sparse TND ($p=0.2$), strong edges are more distributed across hidden-to-hidden pathways, which suggests multi-hop propagation through intermediate neurons before reaching the output. 

Fig.~\ref{fig:act} shows a 2D t-distributed Stochastic Neighbor Embedding (t-SNE) projection of neurons based on their hidden-state trajectories over a 500-timestep window, for TND models of 800 neurons implemented with different $p$. Each point represents one neuron, embedded according to the similarity of its temporal activity pattern to other neurons. At $p=0.2$, neuron activity is broadly scattered, suggesting limited coordination among neurons. As $p$ increases to $0.4-0.6$, neurons form loosely organized groups while still preserving substantial diversity in their temporal dynamics. At $p=0.8$, neurons increasingly synchronize and collapse onto lower-dimensional manifolds, indicating a loss of representational diversity that coincides with degraded performance.

\smallskip
\noindent\underline{\textbf{\emph{Takeaway}}}. The main takeaways are summarized as follows.
\begin{itemize}[leftmargin=*, noitemsep]
    \item Increasing $p$ progressively collapses neuronal activity from distributed patterns onto low-dimensional manifolds, reducing representational capacity. Optimal performance at intermediate sparsity reflects a balance between structured coordination and the heterogeneous temporal dynamics necessary for expressive sequence modeling.
    \item Intermediate-to-sparse connectivity is more suitable for continuous control tasks such as Pong because it promotes hidden-to-hidden multi-hop propagation rather than direct input-to-hidden shortcuts. Such cascaded first-order neuron updates can induce richer higher-order network-level dynamics.
\end{itemize}

\begin{figure}[t]
\centering
\begin{subfigure}{0.48\linewidth}
    \centering
    \includegraphics[width=\linewidth]{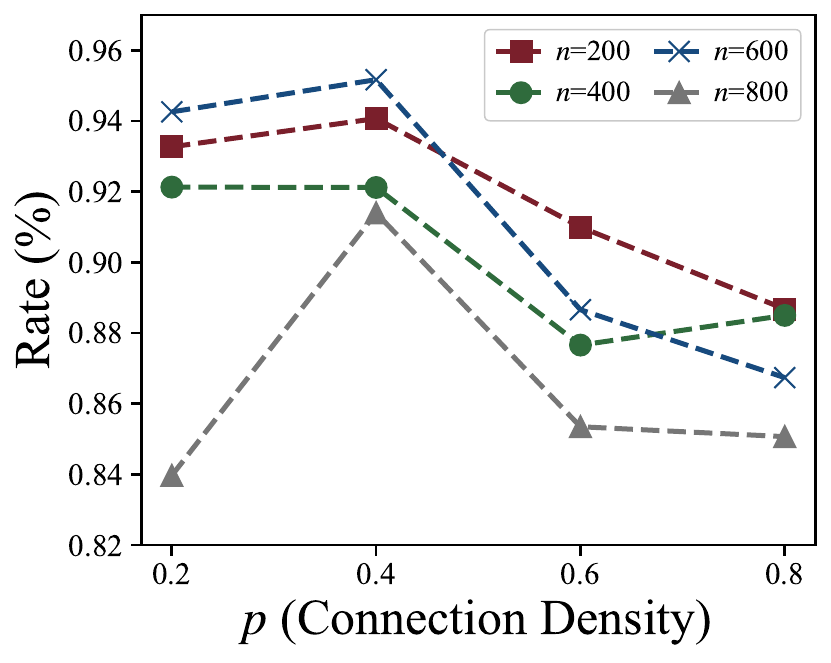}
    \caption{$\mathrm{Rate}$ w.r.t. varying $p$}
    \label{fig:p_rate}
\end{subfigure}
\hspace{-0.5em}
\begin{subfigure}{0.48\linewidth}
    \centering
    \includegraphics[width=\linewidth]{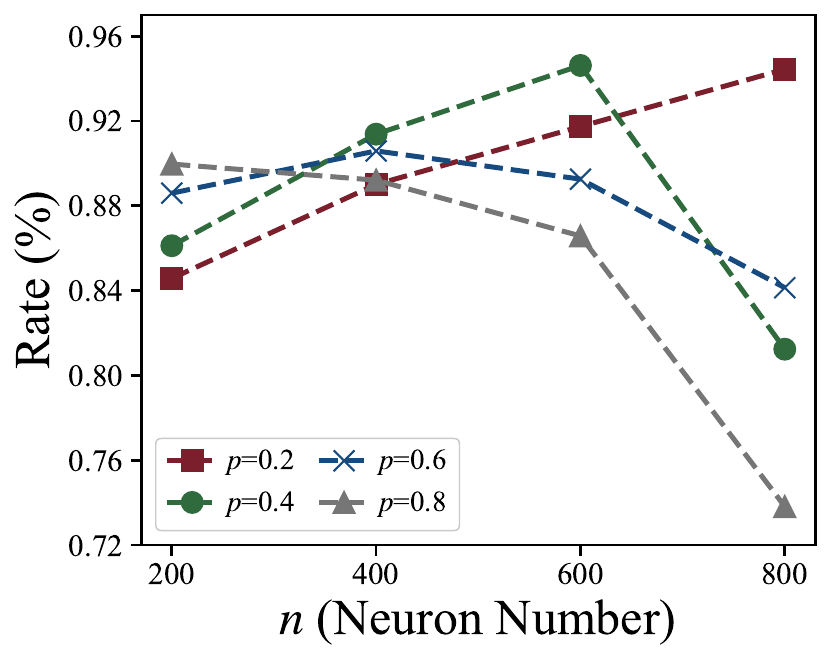}
    \caption{$\mathrm{Rate}$ w.r.t. varying $n$}
    \label{fig:n_rate}
\end{subfigure}
\caption{Parameter analysis.}
\label{fig:para}
\vspace{-1 em}
\end{figure}

\smallskip
\noindent\textbf{TND Parameter Analysis}. 
Fig.~\ref{fig:para} evaluates the impact of two key TND parameters on hit rate: connection density $p$ and neuron number $n$. The results show that TND is sensitive to both topology and model scale. For connection density $p$ (Fig.~\ref{fig:para}(a)), performance is generally higher around $p=0.4$, while overly sparse ($p=0.2$) or overly dense ($p=0.8$) connectivity often leads to lower hit rates. This suggests that TND benefits from an intermediate connectivity level that enables sufficient information propagation without excessive coupling among neurons. For neuron number $n$ (Fig.~\ref{fig:para}(b)), increasing $n$ does not always improve performance. The best results are achieved with moderate neuron numbers, especially around $n=400$--$600$, whereas $n=800$ often degrades performance. This indicates that excessive model size may introduce redundant or unstable dynamics.

\smallskip
\noindent\underline{\textbf{\emph{Takeaway}}}. The takeaways are summarized as follows.
\begin{itemize}[leftmargin=*, noitemsep]
    \item Intermediate connection density, especially around $p=0.4$, achieves better performance in our case, which suggests that TND benefits from a balance between information flow and dynamic independence.
    \item Increasing the number of neurons $n$ improves performance only up to a moderate scale; overly large settings, such as $n=800$, may degrade performance. This indicates that structured topology is more important than simply increasing model size.
\end{itemize}

\begin{table}[t]
\small
\centering
\caption{Ablation Study on Memory Components in TND}
\begin{tabular}{l|ccccccccc}
\toprule
 &State & Recurrent &$\mathrm{Rate}$ &$\mathrm{Mean}$ &$\mathrm{Max}$\\
\midrule
TND$_{\rm no\_rec}$ &$\checkmark$	&$\times$	   &0.28	&0.39	&2 \\
TND$_{\rm no\_state}$   &$\times$    	&$\checkmark$  &0.87	&6.58	&36\\
\textbf{TND}   &\cmark	&\cmark  &\textbf{0.95}	&\textbf{17.47}	&\textbf{68}\\
\bottomrule
\end{tabular}
\label{tab:ab}
\vspace{-1 em}
\end{table}

\smallskip
\noindent\textbf{TND Memory Mechanism Analysis}. 
Table~\ref{tab:ab} evaluates two key memory components of TND for sequential modeling: recurrent connectivity and neuron state. 
We adopt TND$_{\rm no\_rec}$ removes recurrent connections and keeps only feedforward paths, while TND$_{\rm no\_state}$ removes the neuron state $h^t$ by resetting it to zero at each timestep. 
The number of neurons is fixed as 600 (the same as optimal results obtained from TND), and $p$ and $\tau$ are searched the same way. The results show that removing recurrent connectivity causes the largest degradation: TND$_{\rm no\_rec}$ achieves only a 0.28 hit rate and a maximum of 2 consecutive catches in a single round. Removing neuron state is less damaging but still weakens performance, with TND$_{\rm no\_state}$ achieving a 0.87 hit rate and a maximum of 36 consecutive catches in a single round. Full TND performs best across all metrics.

\smallskip
\noindent\underline{\textbf{\emph{Takeaway}}}. There are several takeaways.
\begin{itemize}[leftmargin=*, noitemsep]
    \item Recurrent connectivity is the primary source of memory for TND because it enables information to circulate through the graph across timesteps. Removing it turns TND into a nearly feedforward controller.
    \item Neuron state provides local temporal integration, helping each neuron retain short-term history, but it cannot replace graph-level recurrent propagation.
    \item The full TND performs best because it combines network-level memory from recurrent topology with unit-level memory from neuron dynamics, together forming a richer memory mechanism for sequence modeling.
\end{itemize}

\section{Discussion}
\noindent\textbf{Key Advantage of TND.} The key advantage of TND lies in its shift from layer-wise to neuron-wise dynamics. Conventional sequence models update all neurons through a shared parameterized operator, coupling their temporal evolution and limiting individual flexibility. In contrast, TND treats each neuron as an independent dynamical unit interacting with others through an explicit graph topology, allowing collective computation to emerge from local interactions rather than being imposed by a global operator. Empirically, we observe that this design leads to smoother hidden state trajectories and better performance on the behavior cloning task.

\smallskip
\noindent\textbf{Limitations and Future Work.} We propose TND, a flexible neuron-wise dynamical framework, and demonstrate its effectiveness on a Pong game in the behavior cloning setup. However, several limitations remain. First, the performance of TND can be influenced by the choice of neuron topology, as different tasks may require different interaction structures. Future work could explore data-driven topology learning, sparse graph regularization, and neuronal plasticity, where neurons dynamically adapt their interaction strengths over time, to improve scalability and adaptability. Second, the current instantiation assumes a fixed dynamical function shared across all neurons, which may limit the network's computational diversity. Future work could explore heterogeneous neuron dynamics, where different neurons adopt different dynamical functions, allowing the network to support richer computational repertoires within a single model. Finally, extending TND to broader domains such as biological signal analysis, epidemic modeling, and robotic control may further demonstrate the potential of neuron-wise dynamics as a general framework for complex temporal systems.

\bibstyle{aaai2027}
\bibliography{ref/ref}

\end{document}